\documentclass[journal]{IEEEtran}
\usepackage{cite}

\ifCLASSINFOpdf
\else
\fi
\usepackage{amsmath}
\usepackage{fixltx2e}

\usepackage{stfloats}
\usepackage{comment}
\usepackage{graphicx}
\usepackage{multirow}
\usepackage{float}
\usepackage{wrapfig}
\usepackage{pifont}
\newcommand{\cmark}{\ding{51}}%
\newcommand{\xmark}{\ding{55}}%
\newcommand{\etal}{\textit{et al.}}
\usepackage{siunitx}
\usepackage[dvipsnames]{xcolor}

\begin{document}
%
\title{A Convolutional Baseline for Person Re-Identification Using Vision and Language Descriptions}

\author{Ammarah~Farooq,~
        Muhammad~Awais,~
        Fei~Yan,~
        Josef~Kittler,~
        Ali~Akbari,~
        and~Syed~Safwan~Khalid
\thanks{This work was supported in part by the EPSRC Programme Grant (FACER2VM) EP/N007743/1 and the EPSRC/dstl/MURI project EP/R018456/1.}}

%
%

\markboth{Journal of \LaTeX\ Class Files,~Vol.~14, No.~8, August~2015}%
{Shell \MakeLowercase{\textit{et al.}}: Bare Demo of IEEEtran.cls for IEEE Journals}
%



\maketitle
\begin{abstract}
Classical person re-identification approaches assume that a person of interest has appeared across different cameras and can be queried by one of the existing images. However, in real-world surveillance scenarios, frequently no visual information will be available about the queried person. In such scenarios, a natural language description of the person by a witness will provide the only source of information for retrieval. In this work, person re-identification using both vision and language information is addressed under all possible gallery and query scenarios. 
A two stream deep convolutional neural network framework supervised by cross entropy loss is presented. The weights connecting the second last layer to the last layer with class probabilities, i.e., logits of softmax layer are shared in both networks.
Canonical Correlation Analysis is performed to enhance the correlation between the two modalities in a joint latent embedding space. To investigate the benefits of the proposed approach, a new testing protocol under a multi modal ReID setting is proposed for the test split of the CUHK-PEDES and CUHK-SYSU benchmarks. The experimental results verify the merits of the proposed system. The learnt visual representations are more robust and perform 22\% better during retrieval as compared to a single modality system. The retrieval with a multi modal query greatly enhances the re-identification capability of the system quantitatively as well as qualitatively.
\end{abstract}

\begin{IEEEkeywords}
Cross modal retrieval, vision, language, person ReID, person search.
\end{IEEEkeywords}

%
\IEEEpeerreviewmaketitle

\section{Introduction}
\label{sec:introduction}
\begin{figure}
\centering
        \includegraphics[width=\linewidth]{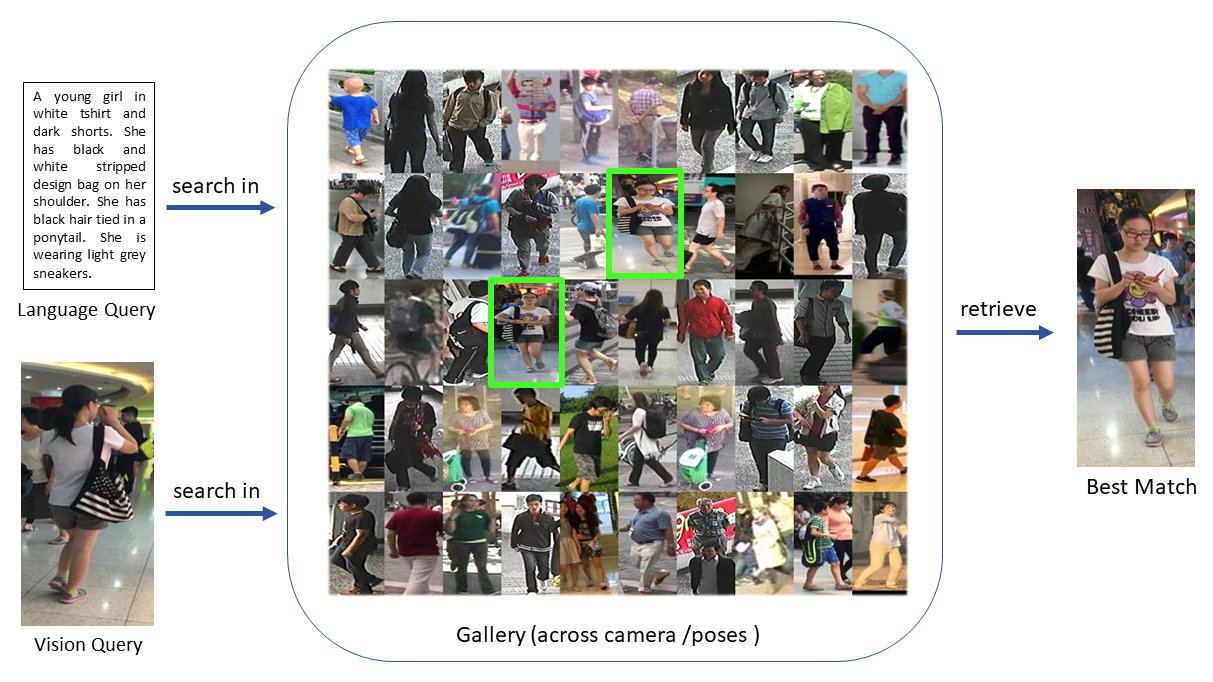}
        \caption{A typical person ReID task takes vision based query only. The proposed system works with vision only, language only as well as both modalities as query.}
    \label{fig:typical}
\end{figure}

%
%
%
%
\IEEEPARstart{I}{n} recent years, the task of person re-identification (ReID) has gained considerable attention in the research community. It aims at recognising a person across non-overlapping camera views. The problem incorporates many underlying computer vision challenges such as illumination changes, occlusion, background changes, pose variations and varying camera resolution. It has significant applications in security and video surveillance, making it even more challenging to work with low resolution CCTV footage. The increasing public safety demands and large networks of installed surveillance cameras are making it difficult to rely solely on manual practice of tracking and spotting a person across various cameras. A typical ReID system takes an image query and searches for the corresponding person in the pool of gallery images (Figure \ref{fig:typical}) or videos. Thus, this scenario assumes that a visual example of a person identity is always available as a query. A huge amount of literature and benchmarking datasets are available for vision based re-identification~\cite{matsukawa2016hierarchical,pedagadi2013local,zhao2013person,liao2015person,mignon2012pcca,koestinger2012large,zhao2017spindle,su2017pose,chen2018improving,xu2018attention,tian2018eliminating}.
Conventional person ReID approaches mainly focus on hand-crafted visual features~\cite{matsukawa2016hierarchical,pedagadi2013local,zhao2013person} and learning discriminative metrics ~\cite{liao2015person,mignon2012pcca,koestinger2012large}. Inspired by the success of convolutional neural networks (CNNs) in large scale visual classification~\cite{russakovsky2015imagenet}, modern approaches~\cite{zhao2017spindle,su2017pose,chen2018improving,xu2018attention,tian2018eliminating} are mainly relying on CNN based feature representations and attention learning mechanisms.

However, in many practical cases, no visual information for  the person of interest is available for matching. In such cases, language description of the person plays an important role to gather useful clues to assist in identification. Such cases are more relevant to the open-set scenarios. For example, we may have a description of a missing person, a description of a criminal by witnesses and shortlisting suspects using appearance cues, where query identity may or may not be present in the available gallery data. Presumably, like images, textual data also contain unique representations to identify a person. Recently, Yan~\etal~\cite{yan2018person} and Li~\etal~\cite{li2017a} pioneered the language based person retrieval, where the former one specifically dealt with person ReID. However, this aspect of person ReID has not yet been explored in more detail by the research community.

Attributes based person retrieval~\cite{vaquero2009attribute, layne2014attributes,su2018multi} offers an alternative to a free-form description based approaches. Such approaches are being used by police where semantic attributes related to clothes, gender, appearance etc. are assigned according to the information given by witnesses and then the search is performed by comparing with the meta-data of the gallery database. However, the use of free-form natural language descriptions offers much more flexibility and robustness as compared to these predefined attributes. Attributes have limited capability of describing persons’ appearance. For instance, the PETA dataset~\cite{deng2014pedestrian} defined 61 binary and 4 multiclass person attributes, while there are hundreds of words for describing a person$'$s appearance. On the other hand, even with a constrained set of attributes, labelling them for a large-scale person image dataset is cumbersome and requires more mindfulness from annotators.

\begin{figure}[t]
    \centering
        \includegraphics[width=\linewidth]{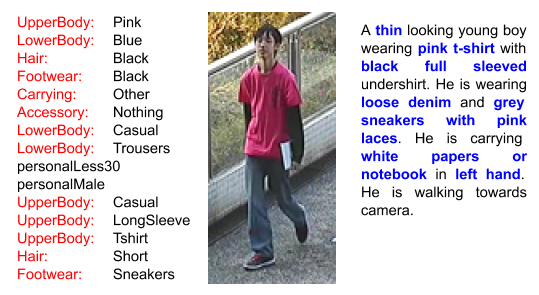}
        \caption{An example of attribute based annotation from PETA dataset vs natural language description}
    \label{fig:peta}
\end{figure}

In contrast to attributes, unique details can be captured by natural language descriptions as shown in Figure \ref{fig:peta}. For example, ``grey sneakers with pink laces" compared to only ``black" and ``sneakers" is more distinctive and accurate. Furthermore, descriptions can include indications of a degree of certitude or ambiguities to preserve as much information as possible (``white papers or notebook''). Another important aspect is the expansion of the set of allowed assignments for attributes (``pink t-shirt with black undershirt" compared to ``pink t-shirt"). Natural language descriptions can describe all such attributes with any possible word and in any possible length. 
Another limitation of the attribute based person retrieval is the sensitivity of the retrieval results. We show in section~\ref{sec:nlp/attributes} that even flipping a few attributes, while using the ground truth set for retrieval, leads to significant performance drop in the results. In practice at test time these attributes are generated automatically, using a machine learning algorithm which will have some error. The sensitive nature of the attribute based person retrieval will have significant impact on the performance in the presence of automatically generated attributes.

With the recent advances in the field of computer vision and natural language processing, joint modelling of vision and natural language is finding applications in image caption generation~\cite{xu2015show,johnson2016densecap,anderson2018bottom}, bidirectional image-text retrieval~\cite{yan2015deep,zheng2017dual,lee2018stacked,gu2018look}, visual question answering~\cite{yang2016stacked,anderson2018bottom}, text to image generation~\cite{zhang2017stackgan,xu2018attngan} and language assisted visual navigation~\cite{chen2019touchdown}. Hence, there is a strong basis to assess the merit of the combined vision and language based techniques in the context of person ReID.

In this work, person ReID using both person images (vision) and corresponding descriptions (language) is addressed. The proposed framework is based on two deep residual CNNs jointly optimised with cross entropy loss to embed the two modalities into a joint feature space. A well-known statistical technique, canonical correlation analysis (CCA) is adopted to further enhance the cross modal retrieval capability. The experimental results on a large scale datasets in various practical scenarios are reported. The scenarios include retrieving from vision to vision, language to vision, vision-language to vision and other combined cross modal scenarios. It extends our conference version paper~\cite{yan2018person} in the following aspects:
\begin{itemize}
\item We propose to change the language network to deep residual network having similar number of layers as the vision branch.
\item We propose a joint optimisation strategy based on cross entropy loss for the two networks along with Canonical Correlation analysis (CCA) to model the embedding space for cross modal retrieval.
\item We propose new testing protocols under the cross modal ReID setting for two large-scale datasets namely CUHK-PEDES and CUHK-SYSU, as compared to Viper and CUHK03 data, to verify the robustness of the proposed approach on a large number of identities coming from diverse domains.
\item The proposed approach is directly compared with state-of-the-art algorithms for person search on the CUHK-PEDES benchmark.
\item Extensive experiments have been performed with large scale data under various practical scenarios: vision to vision, language to vision, language to language and vision language to vision.
\end{itemize}

The rest of the paper is organised as follows. Section \ref{sec:related work} presents a review of existing ReID methods and cross modal retrieval research. Methodology Section \ref{sec:methodology} describes the proposed joint vision-language CNN framework in detail. The implementation details and evaluation protocols are provided in Section \ref{sec:experiments}. In Section \ref{sec:Results}, a discussion on various issues, including training strategies, the experimental results, the merits of natural language descriptions versus attributes is presented along with the qualitative results. Finally, Section \ref{sec:conclusion} concludes the paper.

\section{Related Work}
\label{sec:related work}
\subsection{Vision based Person Re-identification}
Vision based person ReID has gained notable attention in all of its aspects. Research has focused mainly on two issues: robust feature extraction and distance metric learning. The early approaches were focused on improving hand-crafted features such as SIFT, LBP, histograms in colour spaces~\cite{pedagadi2013local} and the hierarchical Gaussian descriptor~\cite{matsukawa2016hierarchical}. Following the success of convolutional neural networks, recent approaches attempt to learn deep representations along with background noise removal~\cite{tian2018eliminating}, using human pose masks~\cite{song2018mask,xiao2016learning,su2017pose} to attend solely people in images and minimize the effect of occlusion. 
Other techniques~\cite{cheng2016person,zhao2017spindle,sun2018beyond,yao2019deep,li2017learning,xu2018attention} focus on fusing global features from the whole image and body part based local features. 

In terms of distance metric learning, Liao~\etal~\cite{liao2015person} proposed cross-view quadratic discriminant analysis (XQDA) that became a popular metric for person ReID. Given features from two views of a person, a low dimensional target space is learnt using eigen decomposition along with a distance function to measure similarity in the subspace. Another measure was proposed by Zhong \etal ~\cite{zhong2017re}, namely re-ranking to use a k-reciprocal nearest neighbours encoding to re-rank the retrieved results using the Jaccard distance to enhance the retrieval performance.

One important work to mention is Chen~\etal~\cite{chen2018improving} who used language as an additional supervision signal in the training phase to learn robust visual features. A global-local strategy was adopted to semantically align the visual features according to text. A global association was established according to the identity labels, while the local association was based upon the implicit correspondences between image regions and noun phrases. However, the retrieval was performed using only vision features. 

\subsection{Person Search}

The task of person search shares the idea of using natural language descriptions for retrieving a person/pedestrian image. However, unlike person ReID, it does not impose constraints of across camera/pose retrieval. The initial work by Li~\etal~\cite{li2017a} introduced a large scale person search (CUHK-PEDES) data. They proposed a CNN-RNN based network to learn word by word affinity of a sentence with image features. The final affinity was obtained by applying a word level gated neural attention mechanism. In a follow-up work, Li~\etal~\cite{li2017b} used a two stage CNN-LSTM network to first learn the identity aware cross modal embedding space using cross modal cross entropy (CMCE) loss. In the next stage, image regions corresponding to each word were spatially attended followed by a latent alignment of the encoded features to enhance the robustness against sentence structure variations. In order to achieve better local matching between two modalities, Chen~\etal~\cite{chen2018adaptive} used a patch-word matching and learnt an adaptive threshold for each word to enhance or suppress its effect on the final text-image affinity. A major improvement in performance was achieved by the work of Zheng~\etal~\cite{zheng2017dual}. They proposed a dual CNN architecture to extract image-text features and used each image-sentence pair as a single class to train the model. Finally, they fine-tuned the networks combined with ranking loss to improve the discriminative ability in a joint embedding space. In a recent work, Jing~\etal~\cite{jing2018cascade} used body pose confidence maps to attend visual features of pedestrians and applied an adaptive similarity based attention mechanism to consider only a description-related part of the image for the final affinity scores.

\subsection{Image-Text Retrieval}
Cross modal person ReID is a fine-grained application of learning a joint image-text embedding for person retrieval in a bi-directional manner. Learning a cross modal embedding has many applications in image captioning~\cite{xu2015show,johnson2016densecap,anderson2018bottom}, visual question answering~\cite{yang2016stacked,anderson2018bottom} and image-text retrieval~\cite{yan2015deep,zheng2017dual,lee2018stacked,gu2018look}.
Recently, in \cite{lee2018stacked}, the authors used a stacked cross attention mechanism to learn the semantic alignment between the objects in image and the corresponding words in the sentence. For each salient object in the image, similarity based attention was computed over the words in the sentence. A relevance score between the corresponding image patch and attended sentence was then computed and used to calculate the final Log-SumExp pooling (LSE) score for matching. A similar attention stacking is applied from text to image and the triplet ranking loss was used as a final objective for optimization. Another work~\cite{gu2018look}, incorporated image-to-text and text-to-image generative models into the conventional cross modal feature embedding. For a given text-input, a GAN was trained to generate an image conditioned on text and similarly, a caption sentence was obtained for the query image. A final representation of each modality is a combination of original features and the features from generated space. Both of these works demonstrated a clear performance gain on MSCOCO and Flicker benchmarks. 
All of these approaches inspire to push the boundary of cross modal matching in the application of person ReID.

\section{Methodology}
\label{sec:methodology}
This section introduces a detailed formulation of the cross modal re-identification system. The cross modal ReID framework, presented in Figure \ref{fig:blockdiagram}, consists of two stream convolutional neural network with identity level classification. This unified framework jointly optimises vision and language modalities to learn a highly distinctive combined feature space. Features obtained from this learnt space are maximally correlated by applying the canonical correlation technique. At the retrieval stage, gallery entities are ranked with respect to their feature similarity with the query feature. Each of the components is discussed in the subsequent sub-sections.

\subsection{Network Architecture}
\subsubsection{Deep Vision Net}
In Figure \ref{fig:blockdiagram}, the upper branch constitutes vision CNN which is based on the ResNet-50~\cite{he2016deep} model. Unlike the original architecture, the proposed network contains two fully-connected (FC) layers before the classifier layer. Batch normalisation is applied after both FC layers and a dropout is applied before the last layer. The network takes input image of size $224\times224$  and generates a 2048 dimensional feature vector $f_{img}$. All images are mean normalized, randomly cropped and horizontally flipped (50\% probability) before passing to the network. 

\subsubsection{Deep Language Net}
\begin{figure*}
    \centering
        \includegraphics[width=\textwidth]{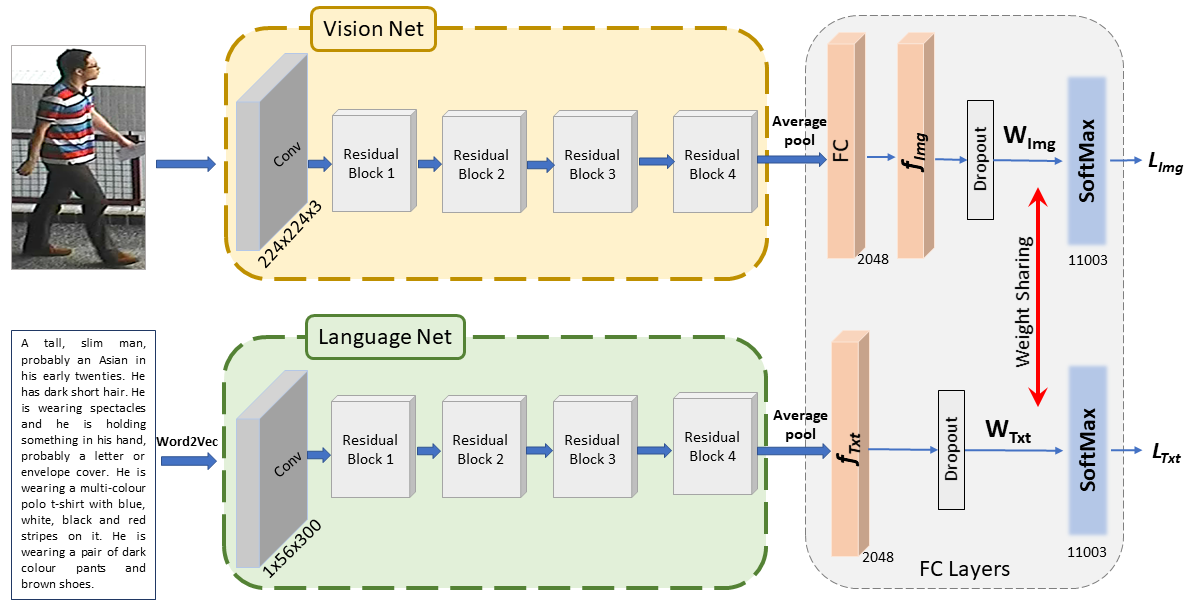}
        \caption{Two stream CNN framework for cross modal person ReID. Feature representations are learnt based on ResNet-50~\cite{he2016deep} model with filter size of $3 \times 3$ and $1 \times 3$ for image and textual inputs respectively. $\boldsymbol{W_{img} = W_{txt} = W_{joint}}$ are shared between two networks during training.}
    \label{fig:blockdiagram}
\end{figure*}

A CNN based language model has been adopted for textual feature learning.
The lower branch of the framework in Figure \ref{fig:blockdiagram} corresponds to the language CNN. It is also a 50 layered deep ResNet~\cite{he2016deep} modified to deal with one dimensional textual input. The first layer is a special word embedding layer which maps each word to a vector space. The rest of the network is similar to the vision net except that the convolution filters are of size $1\times3$ instead of $3\times3$. For each word in a sentence, filter would produce response by looking at the two neighbouring words as well. For a given textual description, the 2048 dimensional feature vector $f_{txt}$ is obtained directly after global average pooling. It is used as a sentence representation at retrieval time. 

\textbf{Text pre-processing and data augmentation:} 
The first step of natural language processing is to map the text corpus into a vector space. Each word in the dictionary is converted to a fixed length vector by an embedding model. In this work, word2vec models pre-trained on Google News corpus (3 billion words) have been used to prepare the dictionary for person descriptions and vector space conversion. Word vectors for words that share common contexts are induced to lie closer to each other in the vector space by using these models. Given a sentence description $S$ containing $n$ words, each word provides a unique index into the corpus. Thus the sentence is converted to a $n \times 300$ matrix where 300 is the output dimension of the word2vec embedding model. As the length of descriptions varies, we have fixed it to the max limit of 56 words to serve as a fixed size input to the CNN. Extra zeros have been appended at the end of shorter sentences. 

Appropriate data augmentation techniques can also be applied to the textual data to create diversity in the training dataset. Similar to random image cropping, a random word dropping operation has been adopted. The number of words to drop is chosen dynamically. Inspired by ~\cite{zheng2017dual}, position shifting for appended zeros is also applied. Instead of keeping all zeros at the end of sentence, their positions are rotated randomly to the beginning of sentence as well.

\subsection{Objective Function}
Under the category-level supervision, CNNs have shown incredible multi-class discriminative ability~\cite{russakovsky2015imagenet}. Zheng~\etal~\cite{zheng2017person} proposed to view person re-identification as a multi-class classification problem and used a transfer learning approach to learn a ReID model, which is termed as identity discriminative embedding (IDE). Strong inter-identity intra-modal representations can be learnt by using this approach. However, for cross modal person Re-ID, given a person image and description, the aim is to learn a joint feature embedding space for retrieval along with such intra-modal inter-identity discriminative representations. Feature representations of a person image and corresponding descriptions are expected to be closer in this joint space. Similarly, for negative pairs, representations should be further away in the joint cross modal space and in the same modality as well. 

The proposed framework is based on identity based supervised learning for both CNNs, where all images of a person and corresponding descriptions are considered as a single class. Hence, two softmax losses $L_{img}$ and $L_{txt}$ are employed, corresponding to each CNN. Specifically, given a batch of $N$ image-text pairs belonging to $I$ person IDs, the image ID loss is the average of cross-entropy loss for all images in the batch and is given as:
\begin{equation}
L_{img} = - \frac{1}{N} \Sigma_{n=1}^N log\bigg(\frac{exp((\boldsymbol{w_{img}})_{i(n)}^T \; f_{img})}{\sum_{j=1}^{I} exp((\boldsymbol{w_{img}})_j^T \; f_{img})} \bigg) , 
\end{equation}
similarly, for text ID loss:
\begin{equation}
   L_{txt} = - \frac{1}{N} \Sigma_{n=1}^N log\bigg(\frac{exp((\boldsymbol{w_{txt}})_{i(n)}^T \; f_{txt})}{\sum_{j=1}^{I} exp((\boldsymbol{w_{txt}})_j^T \; f_{txt})} \bigg) 
\end{equation}
where $i(n)$ represents the target class ID of the $n-th$ pair and $(\boldsymbol{w_{img}})_j$ and $(\boldsymbol{w_{txt}})_j$ are the classifier weights for $j-th$ class. Let $\boldsymbol{{W_{img}}}$ and $\boldsymbol{{W_{txt}}}$ are the classifier weight matrices with weight of each class $j$ in a row. To enforce the learnt representations to be in common space, the weights of the final classifier layer are shared across the image and text CNN imposing  $\boldsymbol{W_{img} = W_{txt}}$ and further denoted as $\boldsymbol{W_{joint}}$. In essence, this joint training approach leads to learning a classifier (weights) for each identity (class) reflecting the information conveyed by both modalities simultaneously and pushing the corresponding $f_{img}$ and $f_{txt}$ pairs closer. The optimization is performed over the combined objective function defined as: 
\[ Loss = \lambda_{1} L_{img}  + \lambda_{2} L_{txt} \]
where $ \lambda_{1}$ and $\lambda_{2}$ set the relative weights for the two losses. During back-propagation, $\boldsymbol{W_{joint}}$ is updated with respect to the gradients from both modalities by taking the average of the gradients.
The classifier weights can also be thought of as a template with which the feature vectors $f_{img}$ and $f_{txt}$ align themselves. In the case of separate weight matrices for each modality, the features vectors $f_{img}$ and $f_{txt}$ will be aligned to their respective weight matrices $\boldsymbol{{W_{img}}}$ and $\boldsymbol{{W_{txt}}}$. In contrast, for the case of training with shared weights both feature vectors $f_{img}$ and $f_{txt}$ will try to align themselves with the joint weight matrix $\boldsymbol{W_{joint}}$, hence, bringing them closer in the same feature space.

\subsection{Canonical Correlation Analysis}
Canonical correlation analysis (CCA)~\cite{hotelling1992relations,hardoon2004canonical} is a well-known statistical technique to find a space in which two sets of random variables are  maximally correlated in the form of their linear combinations. Let $X = (\mathbf{x_1, x_2, ..., x_m}) \in {R}^{d_x \times m} $  and $Y = (\mathbf{y_1, y_2, ..., y_m}) \in {R}^{d_y \times m} $ are the two sets of random vectors with their co-variance matrices denoted as $\Sigma_{xx}$, $\Sigma_{yy}$ and cross co-variance as $\Sigma_{xy}$. The aim of CCA is to explain the correlation structure of $X$ and $Y$ in terms of linear combinations $\mathbf{w_x^T}X$ and $\mathbf{w_y^T}Y$. Concretely, for cross modal ReID, we are seeking those combinations $(\mathbf{w_x^*},\mathbf{w_y^*})$ which maximise the correlation between the two modalities:

\begin{equation}
\begin{split} \label{eq1}
 \rho_{max} &=\; \textup{corr}(\mathbf{w_x^{*T}}X,\mathbf{w_y^{*T}}Y) \\[1ex]
 &= \;  \frac{\textup{cov}(\mathbf{w_x^{*T}}X,\mathbf{w_y^{*T}}Y)}{\sqrt{Var(\mathbf{w_x^{*T}}X) \times Var(\mathbf{w_y^{*T}}Y)}} \\[1ex]
 &= \; \frac{\mathbf{w_x^{*T}}\Sigma_{xy}\mathbf{w_y^*}}{\sqrt{\mathbf{w_x^{*T}}\Sigma_{xx}\mathbf{w_x^*}\mathbf{w_y^{*T}}\Sigma_{yy}\mathbf{w_y^*}}}
\end{split}
\end{equation} 

By substituting $\mathbf{w_x} = \Sigma_{xx}^{-1/2} \mathbf{u}$ and  $\mathbf{w_y} = \Sigma_{yy}^{-1/2} \mathbf{v}$, the above equation becomes
\begin{equation}
\begin{split} \label{eq2}
 \rho_{max} &= \; \frac{{\mathbf{u}^T}\Sigma_{xx}^{-1/2}\Sigma_{xy}\Sigma_{yy}^{-1/2}{\mathbf{v}}}{\sqrt{\mathbf{u}^T\mathbf{u}{\mathbf{v}^T\mathbf{v}}}}
\end{split}
\end{equation} 

In practice, it is assured that $\Sigma_{xx}$ and $\Sigma_{yy}$ are non-singular by using regularization. In that case, a singular value decomposition can be used to solve eq. \ref{eq2} as 
\begin{equation} \label{eq3}
    \Sigma_{xx}^{-1/2}\Sigma_{xy}\Sigma_{yy}^{-1/2} = U\Lambda V
\end{equation}
where $\mathbf{u}$ and $\mathbf{v}$ in eq. \ref{eq2} correspond to the first left and right singular vectors in $U$ and $V$ respectively with top singular value in $\Lambda$ equal to $\rho_{max}$ which gives the maximum correlation between $\mathbf{w_x^{T}}X$ and $\mathbf{w_y^{T}}Y$ and this optimum is attained at $(W_x^*,W_y^*) =({\Sigma_{xx}^{-1/2}} U,  {\Sigma_{yy}^{-1/2}} V)$.

In the proposed framework for cross modal ReID, the 2048 dimensional feature vectors $f_{img}$ and $f_{txt}$ from the two modalities serve as two sets of random variables which are jointly correlated in the $(W_{img}^*,W_{txt}^*)$ space by CCA. 
\section{Experiments}
\label{sec:experiments}
\subsection{Datasets}
\subsubsection{CUHK-PEDES Data}
Currently, the CUHK person description (CUHK-PEDES) data~\cite{li2017person}, introduced for the person search task, is the only large scale available data with natural language descriptions. It contains 40,206 images by combining several ReID datasets, for 13003 identities. Each image is annotated by two sentence descriptions giving about 80,440 descriptions in total. The data has been divided into predefined train/val/test splits.There are 3,074 test images with 6,156 descriptions for 1000 identities, 3,078 validation images with 6,158 descriptions and 34,054 training images with 68,126 descriptions for 11003 identities.

\begin{figure*}
\centering
        \includegraphics[width=0.9\textwidth]{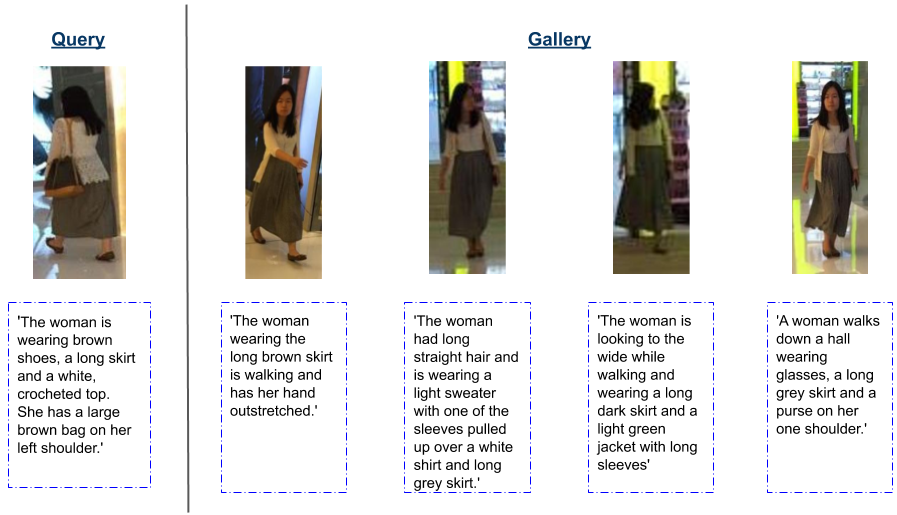}
        \caption{Example of query/gallery pair from annotated cross modal ReID data. Descriptions are annotated per image instead of per identity. Different views have different descriptions and distinctive details.}
    \label{fig:sample}
\end{figure*}

\subsubsection{Cross Modal ReID Data}
Since there is no available data specifically designed for cross modal ReID and to do a fair evaluation under the person ReID constraints, a new evaluation protocol on the test and validation split of CUHK-PEDES has been proposed in this work. The data has been carefully separated across poses and across cameras so that gallery and query images are disjoint (Figure~\ref{fig:sample}). Person IDs having no change in pose have been discarded, resulting in the total of 1594 final distinct IDs. The gallery set contains 2935 images and 5870 descriptions. The query set contains 2127 images and 4264 descriptions. This annotated data is referred to as the cross modal ReID (crossRe-ID) data in the rest of the paper. All the results are reported for the single-shot single query (SS-SQ) scenario.

\subsubsection{CUHK-SYSU Data}
This dataset has been introduced for the joint person detection and identification task~\cite{xiao2016end}. Each identity has at least two images and the descriptions are available from CUHK-PEDES data. The training set contains 15080 images and 30160 descriptions for 5532 identities. The test set includes 8341 images and 16690 descriptions for 2900 identities. A careful annotation has been performed to split poses and viewpoints as much as possible on the test set resulting in 3271 query images and 5070 gallery images. For this dataset, the number of test IDs has been preserved without discarding any image. In most cases where the pose or viewpoint were the same, there were still other challenges like occlusions and severe blur or low resolution. All the results are reported for the SS-SQ scenario.

\subsection{Implementation Details}
The network training has been performed using a stochastic gradient descent (SGD) algorithm with momentum fixed to 0.9. For both CNNs, dropout is set to 0.75 indicating 75\% randomly switched off nodes. The number of FC layers in both networks are empirically chosen to maximise performance. The maximum sentence length is set to 56 following~\cite{zheng2017dual}. The average sentence length is about 19 words. The input to the language net is $1 \times 56$ vector where each entry indicates the index to the word in the dictionary. For the CUHK-PEDES data, dictionary size is 7263 words. The word2vec initialized embedding layer then generates $1 \times 56 \times 300$ dimensional data as input to the first convolution layer. The embedding layer parameters are also tuned during the training to increase the robustness.

The network has been trained for 300 epochs with a batch size 64 and a learning rate of 0.01, which decreases by a multiplicative factor of 0.1 after every 100 epochs. $\lambda_{1}$ and $\lambda_{2}$ are set to 1 for an equal contribution of both modalities. In the next section, the training strategies are discussed, based on the network initialisation choices and the number of learning stages. The implementation has been realised in the PyTorch framework~\cite{paszke2017automatic}.

\subsection{Evaluation} 
Following the ReID literature, three evaluation metrics; recall@K (Rank@1,5,10), mean average precision (mAP) and median rank (medR) have been adopted. At the test time, image feature $f_{img}$ and language feature $f_{txt}$ are obtained independently from the two CNNs. The retrieval is based on cosine similarity between the query and the gallery features which can lie in the range [-1,1]. 
$$S(f_{Query},f_{Gallery})= \bigg(\frac {f_{Query}}{\parallel f_{Query} \parallel}\bigg)^T  \frac {f_{Gallery}}{\parallel f_{Gallery} \parallel}  $$
In the reported results, a query and gallery pair is denoted as Query $\times$ Gallery. Table \ref{table:eval} indicates the gallery and query features used in each retrieval scenario.

\begin{table}[]
\caption{Gallery and query features used for evaluation. $W_{img}^*$ and $W_{txt}^*$ represent CCA projected spaces. [a,b] represents a concatenation of the two vectors.}
\centering
\resizebox{\linewidth}{!}{
\begin{tabular}{c|ccc}
\hline
Retrieval Case & Query Features & Gallery Features & Size\\[0.5ex] \hline  \hline
V X V          & $f_{img}$ & $f_{img}$ & 2048\\[0.5ex]
L X L          & $f_{txt}$ & $f_{txt}$ & 2048\\[0.5ex]
L x V          & ${W_{txt}^*} \times f_{txt}$ & ${W_{img}^*} \times f_{img}$ & 2048            \\[0.5ex] \hline
VL X V         & [$f_{img}$\;, ${W_{txt}^*} \times f_{txt}$] & [$f_{img}$\;, ${W_{img}^*} \times f_{img}$]  & 4096\\[0.5ex]
VL X VL        &[$f_{img}$ \;, $f_{txt}$]  & [$f_{img}$\;, $f_{txt}$]  & 4096   \\ \hline
\end{tabular}}
\label{table:eval}
\end{table}
\section{Results and Discussion}
\label{sec:Results}
\subsection{Training Strategies}
To avoid compromising one modality with another and gain better insight into the joint training of a multi-modal system, the retrieval performance of different training strategies defined by various initialisation of the weights and learning schemes has been analysed. The strategies are presented in Table \ref{table:training}. The weight parameters of the vision CNN and language CNN up to the average pool operation are denoted by $W_{Vision}$ and $W_{Lang}$ respectively. Similarly, the weights for all fully connected layers, including the shared classifier layer, are denoted by $W_{FC}$.We have also trained the vision and language CNNs separately on the CUHK-PEDES data with 11003 train IDs and used these weights for the initialisation of $W_{Vision}$ and $W_{Lang}$. These weights are denoted by ``Person Init'' in the table. For vision, we also used a pre-trained ImageNet model in Strategy 4. In most cases, the networks are trained in two stages, freezing a set of parameters in the first stage and tuning all the parameters together in the second stage. 
\begin{table}[]
\caption{Weights initialisation and learning strategies. $W_{FC}$ are initialised to random values in first training stage and are trained in all stages.}
\centering
\begin{tabular}{c|lccc}
\hline
\multicolumn{1}{l|}{Strategy} & Weights      & \multicolumn{1}{l}{Person Init.} & \multicolumn{1}{l}{ImageNet Init.} & \multicolumn{1}{l}{\begin{tabular}[c]{@{}l@{}}Learnable\\ in Stage 1\end{tabular}} \\ \hline \hline
\multirow{2}{*}{1}            & $W_{Vision}$ & \cmark & \xmark & \xmark\\
                              & $W_{Lang}$   & \cmark &        & \xmark                         \\ \hline
\multirow{2}{*}{2}            & $W_{Vision}$ & \cmark & \xmark & \cmark  \\
                              & $W_{Lang}$   & \cmark &        & \cmark                   \\ \hline
\multirow{2}{*}{3}            & $W_{Vision}$ & \cmark & \xmark & \xmark                     \\
                              & $W_{Lang}$   & \cmark &        & \cmark                      \\ \hline
\multirow{2}{*}{4}            & $W_{Vision}$ & \xmark & \cmark & \xmark                      \\
                              & $W_{Lang}$   & \xmark &        & \cmark                       \\ \hline
\multirow{2}{*}{5}            & $W_{Vision}$ & \xmark & \xmark & \cmark                        \\
                              & $W_{Lang}$   & \cmark &        & \xmark                         \\\hline
\end{tabular}
\label{table:training}
\end{table}

\begin{table}[t]
\caption{Comparison of training strategies on crossRe-ID data}
\centering
 \begin{tabular}{c|  c c c} 
 \hline 
 Rank@1 &  V $\times$ V  & L $\times$ L  & L $\times$ V\\ 
 \hline \hline
St1: pretrained + tune last layers     & 71.2  & 18.38 & 18.69    \\ 
St2: pretrained + tune entirely        & 73.5  & 15.4  & 16.56     \\
St3: pretrained (2 stage training)     & 71.0  & 14.17 & 15.62      \\
St4: ImageNet + Lang. scratch          & 82.05 & 14.93 & 24.6        \\
St5: Vision scratch + pretrained lang. & 71.5  & 19.44 & 18.88        \\[0.5ex] 
 \hline
 \end{tabular}
\label{table:train_results}
\end{table}

Table \ref{table:train_results} compares the performance for the above mentioned strategies. It is interesting to note that ImageNet initialisation (St4) offers better starting point and a wider learning surface as compared to the person specific weights for vision. It is also evident from St3 and St4 where the learning policy is identical. Another observation in this regard is that vision trained from scratch (St5) is similar to that trained from person specific weights (St1,2,3). The choice of initial weights is greatly affecting the results. However, for language, the performance is affected by the number of learning stages. Strategies 1 and 5 performed better with one learning stage for the text CNN as compared to Strategies 3 and 4 where the performance is similar regardless of the initialisation choice. The cross modal retrieval performance is directly proportional to vision; being stronger and less ambiguous compared to language. These results suggest that Strategy 4 be adopted for further evaluation.

\subsection{Results on Cross Re-ID}
In this section, cross modal re-identification performance obtained using the proposed testing protocol is discussed. Since, there is no previous work available for this task, the results are compared with the separate-training technique proposed by Yan~\etal~\cite{yan2018person} in two ways. First, the language network is kept the same as \cite{yan2018person} with one convolution layer after word embedding and two FC layers. This version is referred as ``Yan~\etal~\cite{yan2018person}" in the results. Second, the language network is changed to the proposed deep residual network. It is referred as ``separately train" in the results. In both versions, the vision and language CNNs are trained separately ($\boldsymbol{W_{img} \neq W_{txt}}$) and CCA is applied for joint embedding. The results for the first learning stage of the joint optimisation, where vision CNN is frozen to ImageNet and other network parts of the system are trained from scratch, are also mentioned as Stage 1. Rank@1 comparison is presented in Table \ref{table:crossReid} and the detail quantitative results for each query-gallery scenario are presented in Table \ref{table:detail_crossReid}. 

\begin{table}[]
\caption{Rank@1 retrieval performance on crossRe-ID data}
\centering
\resizebox{\linewidth}{!}{
\begin{tabular}{c|cccc}
\hline
Rank@1 (\%)    & \multicolumn{1}{l}{\begin{tabular}[c]{@{}c@{}}Yan \etal\\~\cite{yan2018person}\end{tabular}} & \begin{tabular}[c]{@{}c@{}}Separately\\ Train + CCA\end{tabular} & \begin{tabular}[c]{@{}c@{}}Jointly Train\\ Stage 1\end{tabular} & \begin{tabular}[c]{@{}c@{}}Jointly Train \\Stage 2 + CCA\end{tabular} \\ \hline
V $\times$ V   & -   & 59.91 & 28.48 & 82.05  \\
L $\times$ L   & 5.39  & 37.32 & 14.8  & 14.93   \\ \hline
L $\times$ V   & 9.22   & 13.6  & 7.46  & 27.9   \\
VL $\times$ V  & 63.6 & 65.87 & 37.7 & 84.7    \\ \hline
VL $\times$ VL & 58.84  & 68.0  & 42.09 & 80.86  \\
VL $\times$ VL & -  & -& -  & 84.06  \\ \hline
\end{tabular}}
\label{table:crossReid}
\end{table}

The results in Table \ref{table:crossReid} validate the scientific hypothesis behind learning the classifier layer jointly. The weights for each ID are forced to achieve maximum alignment between the two modalities. Clearly, the proposed approach outperforms the separate training solutions by a large margin in all cases. Specially, in the cross modal L $\times$ V scenario, the gain is about 14\%. In all other cases, language has helped in learning more distinctive visual features. By using both modalities, the rank@1 performance has further increased by 2.65\%. It indicates that language provides auxiliary information to better define the query ID. Referring to Table \ref{table:detail_crossReid}, Rank@10 suggests that in all scenarios where both modalities are available as a query, the results have the correct match for person in top ten ranks across pose variations with more than 96\% accuracy. However, one striking observation in these results is nearly 4\% decrease in VL $\times$ VL performance as compared to VL $\times$ V. This contrasts with the separate training and Stage 1 for the joint training cases where the increase has been observed. The reason is clear from the severe performance decline of the language modality, while the vision modality appears to become more robust through joint optimisation. We noticed that the language acts as a supervisory signal for vision. Moreover, the vision is a stronger modality, hence after joint training, the shared weights are more influenced by vision and more aligned with it, resulting ina  better performance of VxV. On the other hand the language, being weaker modality, has less influence on the weights and  hence, the performance of LxL drops. The last row of the table corresponds to the performance when we use the jointly trained vision model with the separately trained language model. As expected, a better language model improved the results to 84\% without any joint embedding between the two features. These results offer an interesting direction for further investigation. 

The first two columns of the Table \ref{table:crossReid} support the idea of incorporating a deeper language model. The proposed deep language network achieved higher accuracy as compared to the simple model \cite{yan2018person}. However, for VL $\times$ V, it is interesting to note that even the weaker language model helps in retrieval compared to the vision only scenario.

With regard to the L $\times$ V scenario, note that the above mentioned results correspond to a query sentence description for one view and gallery images corresponding to a different view. From one view to another, the description of person can change greatly and the set of discriminative words to identify the subject across the views may become very small. In such a challenging case, across pose cross modal retrieval accuracy of 27.9\% is quite encouraging. In Table \ref{table:LtoV}, the results are reported for both, within and across pose language to vision retrieval. The joint training not only increased the across pose performance, compared to separately trained networks, but the within pose retrieval is enhanced by a massive 23\%. In comparison to across pose scenario which is more challenging, within pose has higher performance. For example, the performance level for within pose is 4\% better for separately trained networks and 13\% for jointly trained networks. These results broadly support the intuition of cross modal feature alignment with joint optimisation. 

\begin{table}[]
\caption{Detailed retrieval performance on crossRe-ID data in terms of rank@K, mAP and medR}
\centering
\resizebox{\linewidth}{!}{
\begin{tabular}{cccccc}
\hline
\multicolumn{1}{c|}{Model}               & \begin{tabular}[c]{@{}c@{}}Rank@1 \\ (\%)\end{tabular} & \begin{tabular}[c]{@{}c@{}}Rank@5\\ (\%)\end{tabular} & \begin{tabular}[c]{@{}c@{}}Rank@10\\ (\%)\end{tabular} & \begin{tabular}[c]{@{}c@{}}mAP\\ (\%)\end{tabular} & medR \\ \hline \hline
\multicolumn{6}{c}{V $\times$ V}  \\ \hline
\multicolumn{1}{c|}{Separately Train + CCA}    & 59.91 & 80.5 & 85.7 & 64.45 & 1 \\
\multicolumn{1}{c|}{Jointly Train + CCA} & 82.05 & 94.3 & 96.8 & 84.75 & 1  \\ \hline
\multicolumn{6}{c}{L $\times$ V}  \\ \hline
\multicolumn{1}{c|}{Separately Train + CCA}    & 13.6 & 32.99 & 43.04 & 18.5 & 15 \\
\multicolumn{1}{c|}{Jointly Train + CCA} & 27.9 & 50.6  & 60.7  & 33.4 & 5   \\ \hline
\multicolumn{6}{c}{VL $\times$ V} \\ \hline
\multicolumn{1}{c|}{Separately Train + CCA}    & 65.87 & 84.19 & 88.9 & 64.8 & 1 \\
\multicolumn{1}{c|}{Jointly Train + CCA} & 84.7 & 95.0 & 97.1 & 84.1 & 1    \\ \hline
\multicolumn{6}{c}{VL $\times$ VL} \\ \hline
\multicolumn{1}{c|}{Separately Train + CCA}    & 68.0 & 84.7 & 89.58 & 71.8 & 1    \\
\multicolumn{1}{c|}{Jointly Train + CCA} & 80.86 & 94.16 & 96.6 & 83.85 & 1   \\ \hline
\end{tabular}}
\label{table:detail_crossReid}
\end{table}

\begin{table}[]
\caption{Text-to-Image retrieval performance on crossRe-ID data within pose and across pose}
\centering
\begin{tabular}{l|ccc}
\hline
\multicolumn{1}{c|}{\begin{tabular}[c]{@{}c@{}}L x V\\ Rank (\%)\end{tabular}} & \begin{tabular}[c]{@{}c@{}}Separately\\ Train + CCA\end{tabular} & \begin{tabular}[c]{@{}c@{}}Jointly Train\\ Stage 1\end{tabular} & \begin{tabular}[c]{@{}c@{}}Jointly Train\\ stage 2 + CCA\end{tabular} \\ \hline \hline
Across Pose & 13.4 & 7.46  & 27.9 \\
Within Pose & 17.6 & 10.8 & 40.46 \\
\hline                       
\end{tabular}
 
\label{table:LtoV}
\end{table}

\subsection{Results on Person Search}
\begin{figure*}[]
\centering
        \includegraphics[width=0.9\linewidth, height = 0.5\linewidth]{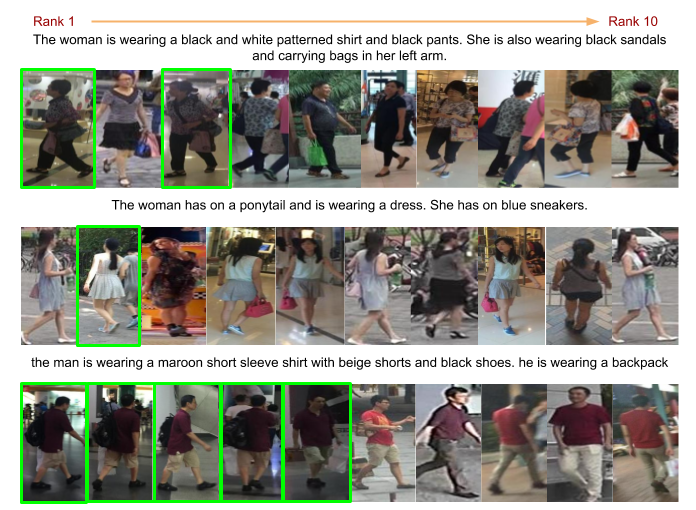}
        \caption{Qualitative results on CUHK-PEDES in the LXV scenario. Correct matches are shown by green boxes.}
    \label{fig:pedes}
\end{figure*}

The proposed approach has been evaluated on the person search task under L $\times$ V scenario. The quantitative and qualitative results on the test set of CUHK-PEDES are reported in Table~\ref{table:pedes} and Figure~\ref{fig:pedes}, respectively. The results are compared with the current state-of-the-art DPC approach~\cite{zheng2017dual} and other techniques. Note that Zheng~\etal~\cite{zheng2017dual} has also used a two stage training procedure by including a ranking loss along with ID level losses in the second stage. The proposed joint learning with CCA and identity loss only has achieved competitive results with 2\% boost in rank@1 accuracy. 

\begin{table}[h!]
\caption{Retrieval performance on the CUHK-PEDES data}
\centering
\resizebox{\linewidth}{!}{
\begin{tabular}{c|ccccc}
\hline
Model & \begin{tabular}[c]{@{}c@{}}Rank@1\\ (\%)\end{tabular} & \begin{tabular}[c]{@{}c@{}}Rank@5\\ (\%)\end{tabular} & \begin{tabular}[c]{@{}c@{}}Rank@10\\ (\%)\end{tabular} & \begin{tabular}[c]{@{}c@{}}mAP\\ (\%)\end{tabular} & medR      \\ \hline \hline
Yan et al. ~\cite{yan2018person}& 17.5   & 35.25 & 46.34  & 16.65 & 13 \\
Separately Train + CCA & 19.6   & 38.88 & 50.2  & 18.2 & 10 \\
GNA-RNN ~\cite{li2017person}    & 19.05  & -  & 53.64 & - & -     \\
IATV ~\cite{li2017identity}     & 25.94  & -  & 60.48 & - & -      \\
GDA ~\cite{chen2018improving}   & 43.58  & 66.93 & 76.26 & - & -    \\
DPC ~\cite{zheng2017dual}       & 44.4   & 66.26 & 75.07 & -  & 2    \\ \hline
\begin{tabular}[c]{@{}c@{}}Jointly train\\ + CCA (proposed)\end{tabular} & 46.44 & 67.9  & 76.3 & 41.3 & 2    \\ \hline
\end{tabular}}
\label{table:pedes}
\end{table}

\subsection{Results on CUHK-SYSU}
The results under the re-identification setting on CUHK-SYSU are presented in Table \ref{table:detail_crossReid_sysu}. As mentioned earlier, the test data is annotated under the strict ReID setting and the challenging single shot setting. Again, vision achieved nearly a  20\% gain by learning jointly with language. Rank@1 of 77.82\% and 80.5\% mean average precision is achieved under the multi modal scenario. The results obtained are quite competitive and set a baseline for future work. The choice of this data for cross modal ReID is mainly due to its size, number of IDs and the availability of corresponding language descriptions.

\begin{table}[h!]
\caption{Detailed retrieval performance on CUHK-SYSU data in terms of rank@K, mAP and medR}
\centering
\resizebox{\linewidth}{!}{
\begin{tabular}{cccccc}
\hline
\multicolumn{1}{c|}{Model}               & \begin{tabular}[c]{@{}c@{}}Rank@1 \\ (\%)\end{tabular} & \begin{tabular}[c]{@{}c@{}}Rank@5\\ (\%)\end{tabular} & \begin{tabular}[c]{@{}c@{}}Rank@10\\ (\%)\end{tabular} & \begin{tabular}[c]{@{}c@{}}mAP\\ (\%)\end{tabular} & medR \\ \hline \hline
\multicolumn{6}{c}{V $\times$ V}  \\ \hline
\multicolumn{1}{c|}{Separately Train + CCA}    & 55.03 & 71.41 & 77.89 & 58.95 & 1  \\
\multicolumn{1}{c|}{Jointly Train + CCA} & 74.13 & 87.2  & 90.48 & 77.16 & 1  \\ \hline
\multicolumn{6}{c}{L $\times$ V}  \\ \hline
\multicolumn{1}{c|}{Yan \etal~\cite{yan2018person}}    & 3.03  & 11.06  & 17.51  & 5.48  & 89 \\
\multicolumn{1}{c|}{Separately Train + CCA}    & 2.31  & 9.13   & 15.06  & 4.57  & 94 \\
\multicolumn{1}{c|}{Jointly Train + CCA} & 11.37 & 28.41  & 38.06  & 15.78 & 24  \\ \hline

\multicolumn{6}{c}{VL $\times$ V}                      \\ \hline
\multicolumn{1}{c|}{Yan \etal~\cite{yan2018person}}    & 58.72 & 74.82 & 79.44 & 57.01 & 1 \\
\multicolumn{1}{c|}{Separately Train + CCA}    & 58.72 & 74.79 & 79.44 & 57.0 & 1 \\
\multicolumn{1}{c|}{Jointly Train + CCA} & 77.68 & 89.2  & 92.03 & 75.8 & 1  \\ \hline

\multicolumn{6}{c}{VL $\times$ VL}                        \\ \hline
\multicolumn{1}{c|}{Yan \etal~\cite{yan2018person}}    & 57.77 & 77.17 & 82.13 & 62.13 & 1  \\
\multicolumn{1}{c|}{Separately Train + CCA}    & 59.65 & 76.27 & 81.34 & 63.3 & 1  \\
\multicolumn{1}{c|}{Jointly Train + CCA} & 77.82 & 89.31 & 92.41 & 80.4 & 1   \\ \hline


\end{tabular}}
\label{table:detail_crossReid_sysu}
\end{table}

\subsection{Natural Language vs Attributes}
\label{sec:nlp/attributes}
The task of person re-identification using descriptions closely resembles the attributes based person retrieval. In this section, we show, with the help of experiments, that using unstructured language descriptions as compared to attributes enhances retrieval robustness. For this purpose, we choose the Market-1501 dataset which has been manually annotated with 27 appearance based attributes~\cite{lin2019improving}. There are 8 colours for upper-body clothing, 9 colours for lower body clothing, 1 for age with four possible values and 9 other binary attributes including gender, hair length, carrying backpack, wearing hat etc. For evaluation, we use 750 test set identities. First we extract their corresponding language descriptions from CUHK-PEDES which contains around four images per identity. We then select from this subset 750 images from one camera as the gallery and 750 images from another camera as a query set. Finally, we concatenate the corresponding attributes or descriptions for retrieval.

Since the attributes are annotated at the identity level, there exists no notion of viewpoint. When we concatenate ideal attributes for gallery and query, 93.46\% rank@1 is achieved. However, in practical systems, it is highly unlikely to obtain all attributes correctly on an unseen query. If attributes from one viewpoint are provided for search, then the system may overlook other important attributes which may be visible from other viewpoint and tries to match according to the given set only. In Table~\ref{table:attribute-exp} we investigate the sensitivity of attribute based retrieval to missing attributes. Note that if we change just one attribute in query (N=1), we observe 4.5\% decrease. By changing two attributes, the performance we achieve drops again and becomes comparable to VL $\times$ VL. The accuracy decreases by 7.2\% compared to sentence descriptions and by 20\% from ideal attributes just by flipping three random attributes. Practical attribute prediction systems can not produce 100\% accurate attribute features for unseen data and consequently the retrieval will always be subject to attribute flipping . The results reported in the table are obtained with ground truth attributes and still a significant decrease in the performance has been observed. With the predicted attribute features, retrieval will always suffer from viewpoint variance. On the other hand, the proposed joint training system is trained on image level annotations with many different words and sentences to describe one person. In such a case, the system tries to learn different semantic concepts present in the images instead of memorising individual identities. 

\subsection{Discussion on Language to Language Performance}
As observed in Table \ref{table:crossReid}, vision features become more discriminative when trained with language descriptions.In contrast language only performs poorly, compared to the separately trained system. One possible reason could be that as the joint optimisation forces the two modalities to lie closer in the embedding space, language tends to become more specific to match with its corresponding image because the weights $\boldsymbol{W_{joint}}$ serve as a common template for alignment. These weights are more influenced by stronger modality and thus force the weaker modality to follow along. As a result, language becomes more sensitive compared to the vision modality.
An evidence that can be qualitatively observed is that after joint training the overall text classification (word-to-word) accuracy has been improved as shown in Table \ref{tab:Lang-qual}. Since the ReID evaluation criterion is based on identity, the L $\times$ L retrieval exhibits a decrease as the same set of words can describe two different persons. For future investigation, it will be interesting to examine adjective-noun association, impact of non-informative frequent words and the organisation of the sentences for learning feature representations.

\begin{table}[]
\caption{Rank@K performance on the Market-1501 data with vision and attributes}
\centering
\resizebox{\linewidth}{!}{
\begin{tabular}{c|ccccc}
\hline
                & \begin{tabular}[c]{@{}c@{}}Rank @ 1\\ (\%)\end{tabular} & \begin{tabular}[c]{@{}c@{}}Rank @ 5\\ (\%)\end{tabular} & \begin{tabular}[c]{@{}c@{}}Rank @ 10\\ (\%)\end{tabular} & \begin{tabular}[c]{@{}c@{}}mAP\\ (\%)\end{tabular} & medR \\ \hline \hline
                
V $\times$ V           & 74.66  & 92.66 & 96.13 & 78.70 & 1   \\
VL $\times$ VL         & 81.60  & 95.46 & 97.46 & 84.64 & 1    \\ \hline
VA $\times$ VA         & 93.46  & 99.4  & 100   & 94.89 & 1     \\
VA $\times$ VA (N = 1) & 88.93  & 99.2  & 100   & 91.2  & 1    \\
VA $\times$ VA (N = 2) & 82.93  & 97.86 & 99.3  & 86.25 & 1     \\
VA $\times$ VA (N = 3) & 74.40  & 95.73 & 98.53 & 78.85 & 1      \\
VA $\times$ VA (N = 4) & 62.12  & 90.26 & 94.9  & 67.98 & 1       \\
VA $\times$ VA (N = 5) & 50.53  & 78.80 & 88.26 & 56.87 & 1        \\ \hline
\end{tabular}}
\label{table:attribute-exp}
\end{table}

\begin{table*}[]
\centering
\caption{Qualitative results for L $\times$ L using jointly trained model. Samples provided are correctly retrieved by separately trained model. The highlighted phrases show the identical words in the query and retrieved sentence.}
\resizebox{\textwidth}{!}{%
\begin{tabular}{lll}
\hline
\multicolumn{1}{c}{Query} & \multicolumn{1}{c}{Ground Truth}   & \multicolumn{1}{c}{Retrieved}     \\
\hline \hline
    \begin{tabular}[c]{@{}l@{}}`A man wearing a black jacket with long sleeves, \\ a pink shirt, a pair of dark colored pants and \\a pair of black shoes.'\end{tabular}                                           & \begin{tabular}[c]{@{}l@{}}`I am watching a man walking by, he is wearing all black\\ with a pink dress shirt, and a long coat, carrying a \\newspaper, and looking around as if he's somewhat lost.'\end{tabular}               & \begin{tabular}[c]{@{}l@{}}`A woman \colorbox{Yellow}{wearing a pink shirt, a pair of} black \colorbox{Yellow}{pants}\\ \colorbox{Yellow}{and a pair of black shoes.}'\end{tabular}  \\ [0.5ex]
    \hline
\begin{tabular}[c]{@{}l@{}}`The lady wears a white and black shirt white\\ and black pants with grey and white sneakers she\\ walks inside the building and carries a black back pack.'\end{tabular} & \begin{tabular}[c]{@{}l@{}}`The woman is wearing a black and white patterned outfit,\\ and has a large pack on her back.'\end{tabular}         & \begin{tabular}[c]{@{}l@{}}`The man \colorbox{Yellow}{wears a} blue and \colorbox{Yellow}{white} stripped \colorbox{Yellow}{shirt black pants}\\ with black and \colorbox{Yellow}{white sneakers} he \colorbox{Yellow}{carries a black back pack.}'\end{tabular}                                            \\ [0.5ex]
\hline
\begin{tabular}[c]{@{}l@{}}`A pedestrian walking to the right with a black bag\\ on their back,grey pants, and red shoes. '\end{tabular}                                                                   & \begin{tabular}[c]{@{}l@{}}`The girl has her hair tied back and has a back pack on.\\  She has her back to the onlooker and has dark pants on. \\ Her pack has a touch of red on it and her sneakers are red as well.'\end{tabular} & \begin{tabular}[c]{@{}l@{}}`This person \colorbox{Yellow}{is walking.} They have on a \colorbox{Yellow}{red} shirt and light shorts.\\ they have a purse or \colorbox{Yellow}{bag over their right} shoulder and\\ are carrying something in their right hand.'\end{tabular} \\  [0.5ex]
\hline 
\begin{tabular}[c]{@{}l@{}}`The woman is wearing a white dress and pink shoes.\\ She is carrying a black bag and is looking down at her phone.'\end{tabular}                                             & \begin{tabular}[c]{@{}l@{}}`white dress asain women pink shoes black long hair wearing\\ backward backpack and glasses black back pack and gold phone'\end{tabular}                                                               & \begin{tabular}[c]{@{}l@{}}`\colorbox{Yellow}{The woman is} talking on the \colorbox{Yellow}{phone} and she \colorbox{Yellow}{is wearing a white}\\ \colorbox{Yellow}{dress} with white \colorbox{Yellow}{shoes.} She also \colorbox{Yellow}{is carrying a} grey and black shoulder \colorbox{Yellow}{bag}'\end{tabular} \\ \hline
\end{tabular}%
}
\label{tab:Lang-qual}
\end{table*}

\subsection{Qualitative Results}

\begin{figure*}[]
\centering
        \includegraphics[width=\linewidth, height=0.4\linewidth]{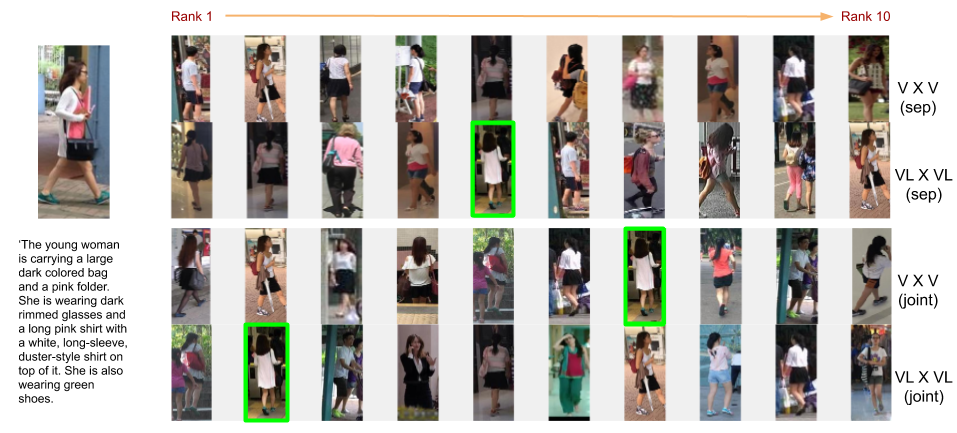}
        \includegraphics[width=\linewidth, height=0.4\linewidth]{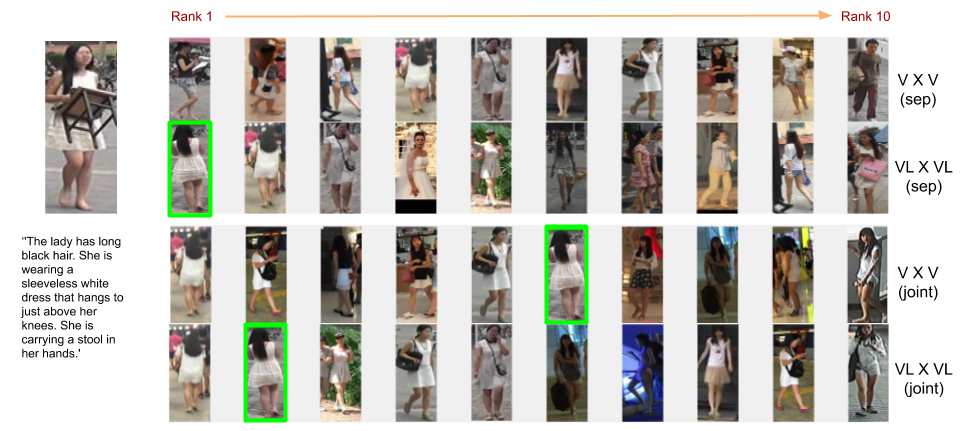}
        \includegraphics[width=\linewidth, height=0.4\linewidth]{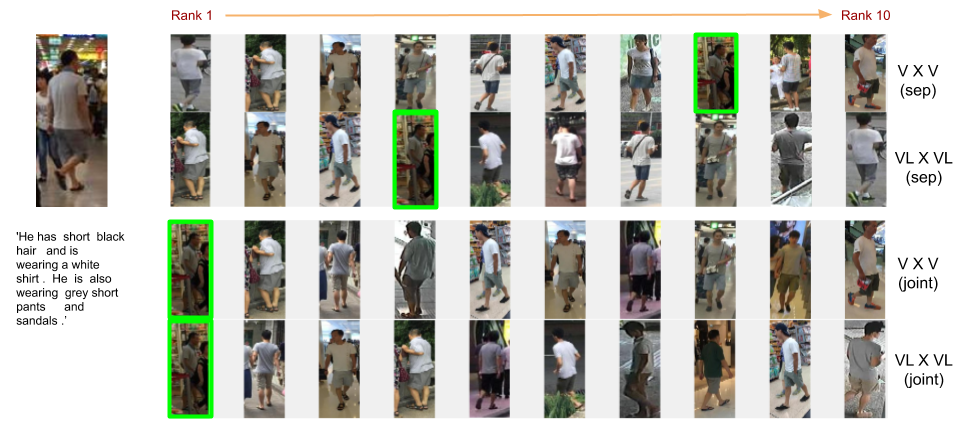}
        \caption{Qualitative results on crossRe-ID dataset. Correct matches are shown by green boxes. For each sample, first two rows show results for separately trained model and last two rows show results for the jointly trained model. For both models, language has refined the feature representations and boosted the retrieval performance.}
    \label{fig:reId1}
\end{figure*}

Figure \ref{fig:reId1} presents qualitative results on the crossRe-ID dataset. The results are shown for both separate and joint modelling. In the first two examples, there is no correct match for the given query for V $\times$ V case with separate modelling. However, it can be seen that with the help of language, correct matches are observed in top five retrieved images. Specifically, if we look at the second example, top match for the vision only (separate training) is influenced by the background of the image. On the other hand, textual description served as an attention to focus on the person only features. Thus, the top five matches contains a white dress and black hair. In case of joint training, it is evident that the corresponding vision features become stronger and use of language feature further pushes the correct match towards the top ranks. The performance of the proposed system has also been tested practically, including using query images from CCTV cameras which are not part of the dataset. This exercise has helped in gaining insight on the challenges in practical applications. We have observed that with image only, we are able to get closer to main concepts like gender of person and colours of the clothes. By adding language description to image, we obtain refined results capturing more details. 

\begin{figure}[h!]
\centering
        \includegraphics[width=\linewidth]{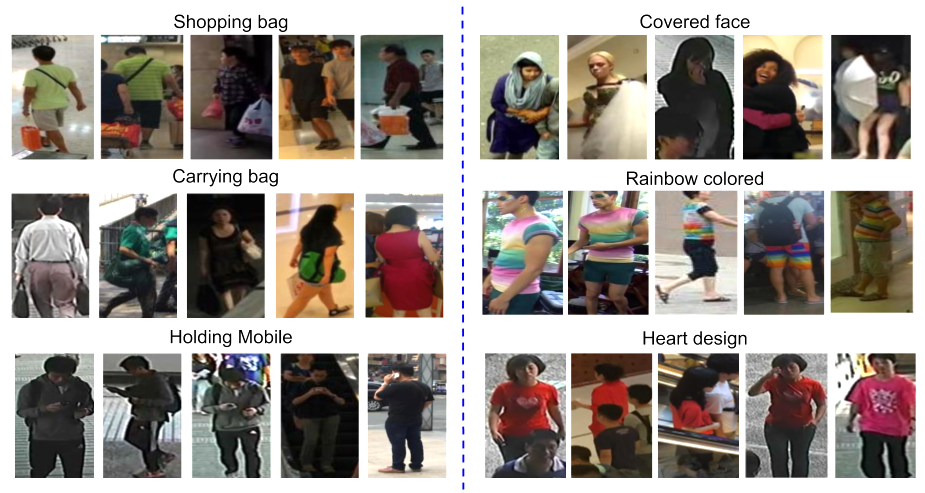}
        \caption{Retrieved images corresponding to different semantic concepts}
    \label{fig:reID2}
\end{figure}

It is also interesting to examine whether the individual semantic concepts in a language description are implicitly represented by the individual visual units in $f_{img}$. Some examples of the images obtained for various semantic concepts are in Figure \ref{fig:reID2}. Instead of the full description, we just passed the desired phrases as input to the language CNN. The results in the figure confirm the presence of a specific visual unit. This kind of search also assists in database shortlisting for a query, for example ``covered face''. One important aspect for future studies from these observations is to check the impact of less frequent words.

\section{Conclusion}
\label{sec:conclusion}
Person re-identification by jointly modelling vision and language reflects better practical scenarios in security applications. In this paper, an integrated baseline framework has been proposed, based on the joint optimisation of the two modalities. We have investigated various training strategies to develop deep understanding of the proposed joint-embedding learning and evaluated the performance on all the possible query-gallery combinations. The proposed joint embedding learning and the two stage training protocol achieved superior results as compared to the separate training strategy, outperforming it by 22\% in V$\times$V, 11\% in L$\times$V, 18\% in VL$\times$V and  12\% in VL$\times$VL on crossRe-ID data. These results lay the groundwork for future research in cross modal Re-ID. The proposed method is quite promising as it is competitive to current state-of-the-art for the person search task. Evidently, language provides complementary information and facilitates learning enriched representations for person images. Using free-form descriptions has provided a significant edge over attribute based retrieval in terms of relaxed annotation constraints, incorporating more unique details and overall more powerful language modelling. 


%








\bibliographystyle{IEEEtran}
%
\bibliography{ebib.bib}


%

\end{document}